# Personalized Patent Claim Generation and Measurement

Jieh-Sheng Lee[1]

Department of Computer Science and Information Engineering
National Taiwan University, Taiwan
`d04922013@csie.ntu.edu.tw`

**Abstract.** This work-in-progress paper proposes a framework to generate and measure personalized patent claims. The objective is to help inventors conceive better inventions by learning from relevant inventors. Patent claim generation is a way of "augmented inventing." for inventors. Such patent claim generation leverages the recent transfer learning in the Deep Learning field, particularly the state-of-the-art Transformer-based models. In terms of system implementation, it is planned to build an "auto-complete" function for patent claim drafting. The auto-complete function is analyzed from four different perspectives: extent of generation, generative direction, proximity of generation, and constraint in generation. Technically, the framework is composed of two Transformer models. One is for text generation and the other is for quality measurement. Specifically, the patent claim generation is based on GPT-2 model and the measurement of personalization is based on BERT model. The training data is inventor-centric and comes from the Inventors Endpoint API provided by the USPTO.

## 1 Introduction

### 1.1 Transfer Learning & Augmented inventing

In the computer science field, NLP (Natural Language Processing) turns text into structured data and NLG (Natural Language Generation) turns structured data back to text. Transfer learning is a method where a model trained for a task (either NLP or NLG) is reused as the starting point for fine-tuning the model on a second task. Recently, transfer learning based on Transformer models [1], such as BERT [2] and GPT-2 [3], has resulted in significant state-of-the-art performances. Such transfer learning is implemented by pre-training an unsupervised language model on a large corpus and fine-tuning the model on downstream tasks with much fewer data. In terms of architecture, a Transformer model comprises an Encoder and a Decoder. The Encoder is capable of turning text into structured data, such as tensors in artificial neural networks, and the Decoder is capable of turning tensors back to text. Both of the Encoder and Decoder can work as a standalone model without the other. For example, BERT and GPT-2 are the most noteworthy Encoder and Decoder respectively. In previous works, I had experimented on a classifier based on BERT [4], a prototype of patent claim generation based on GPT-2 [5], and a framework to measure the text generation of GPT-2 by using BERT [6]. Based on these experiences, this paper moves forward to propose a framework for personalized patent claim generation.

---

[1] Admitted in New York and passed the USPTO patent bar exam. Currently a Ph.D. candidate focusing on Deep Learning for patents and an in-house patent counsel at Novatek Microelectronics Corp.

Transfer learning is a means to an end. The ultimate goal is to build an "augmented inventing" system, so that transfer learning of inventive minds is made possible. For example, it is planned to implement an "auto-complete" function in which, if an inventor is contemplating and has no whole picture in mind yet, patent claim generation can augment the inventor to explore relevant ideas. Such interaction between human and machine will also open a window for both qualitative and quantitative analysis on augmented inventing. By measuring how the inventor responds to the system, it is possible to collect human annotations for active learning. The second part of this paper will explain the auto-complete function in details.

### 1.2 Personalization

It is planned to fine-tune a pre-trained model with inventor-centric data for reaching personalization. For example, the data may start from a seed inventor and expand to include more patents from other inventors through patent citations. The depth of citations decides how many patents to include. Extra keywords may be used to include or exclude patents. By collecting patents from similar minds, it is hypothesized that the fine-tuned GPT-2 model can generate claim text of higher relevancy. Such an inventor-centric approach might be possible because of the PatentsView API[2] provided by the USPTO. The API provides web developers and researchers programmatic access to longitudinal data and metadata on patents, inventors, companies, and geographic locations.[3] Notably, the Inventors Endpoint[4] of the API can search for inventors who had patents granted based on date range, country or city, CPC classification or other criteria. A primary underlying challenge in the API is inventor disambiguation since the USPTO does not require an inventor to record a unique identifier. Searching for all of the patents associated with a specific inventor can be difficult, particularly if the inventor's name has multiple forms. In 2015, the USPTO hosted an Inventor Disambiguation Workshop. The winning team, led by Andrew McCallum and Nicholas Monath [7], uses discriminative hierarchical coreference as a new approach and reached 98.27% in their F1 score [8]. Their algorithm was integrated into the PatentsView data platform in March 2016. Inventor disambiguation is the reason why inventor-centric training data in this work should be feasible. It is noted that the API 4provides seven endpoints in total, such as Assignees Endpoint, Location Endpoint, CPC Endpoint, etc. Therefore, if the personalization implemented in this paper works, the idea can be generalized and applied to different data perspectives, such as generating patent claims which are specific to a company or a city.

---

[2] http://www.patentsview.org
[3] http://www.patentsview.org/api/faqs.html
[4] http://www.patentsview.org/api/inventor.html

### 1.3 A Span-based Approach

In the NLP field, language modelling is the task of predicting what word comes next. Specific to the patent claim language, a span-based approach is to predict what "claim span" comes next. A claim span is a segment of text in a patent claim based on human annotations. For example, claim 1 of US9229634B2 is split into spans as **Fig. 1**. How to split a claim into spans is skipped here for brevity, and interested readers can refer to [5] [6] for details.

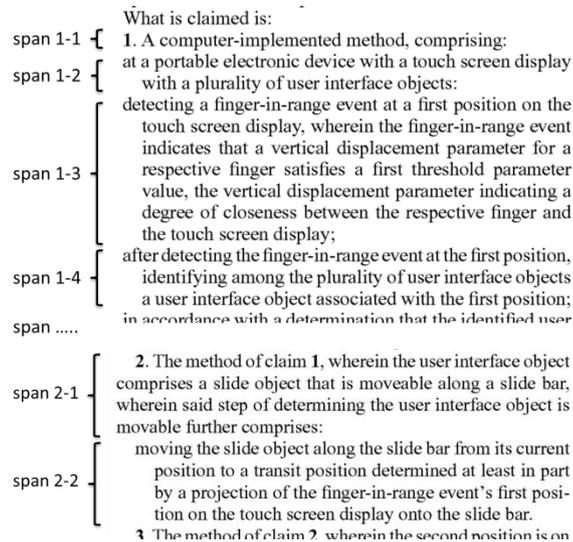

**Fig. 1.** Spans in the '634 patent

A span boundary in a patent claim is made by human and therefore a human annotation. Such a boundary is meaningful for patent readers. In this paper, it is assumed that a Transformer may learn or generate more meaningful data based on claim spans. My previous experiments in [4] [5] [6] showed effective results of the span-based approach. Coincidentally, among the various BERT-like models, Joshi et al. [9] propose the SpanBERT, which is a pre-training method that is designed to better represent and predict spans of text. According to the authors, SpanBERT consistently outperforms BERT, by masking contiguous random spans (rather than random tokens in the original BERT) and training the span boundary representations to predict the entire content of the masked span. For this paper, it is assumed that SpanBERT is further validation of the span idea on Transformer-based models.

### 1.4 Framework

The framework is composed of two Transformers. In the previous work [6], I define a similar framework to experiment with the approach of measuring span relevancy in patent claim generation. In that work, conceptually, it uses a fine-tuned Transformer Encoder to measure a fine-tuned Transformer Decoder. In this paper, the concept is generalized further as **Fig. 2**. On its left-hand side, the first fine-tuned model is based on a pre-trained model. The fine-tuning is based on training data in a specific domain, e.g. patent claims. In order to make text generation personalized, the pre-trained model is further fine-tuned by inventor-centric data, e.g. patents within a certain degree of relevancy to an inventor. By doing so, a hypothesis to validate is that the generated patent claims will be more relevant to the inventor. On the right-hand side, the same two steps of fine-tuning with specific domain data and inventor-centric data are similar, except for working on a different Transformer model.

A metric calculation on the right-hand side is designed to feed measurement results to the text generation on the left-hand side. For example, when the sampling algorithm in text generation produces multiple candidates in its search space, the metric can be used for choosing the best result. In the middle of the framework, the text processing is generalized since conceptually the text can be a token (sub-word), a word, a phrase, a span or a complete sentence. The frequency of interaction between the left-hand side and right-hand side is also generic. It could be real-time or batched. In the previous work, the text processing is span-based while the text generation is word by word. The metric calculation operates in a batch manner after generating a batch of patent claims. In brief, this paper generalizes the elements in the previous framework when possible. The reason for further generalization and higher flexibility is to meet the needs of different auto-complete functions.

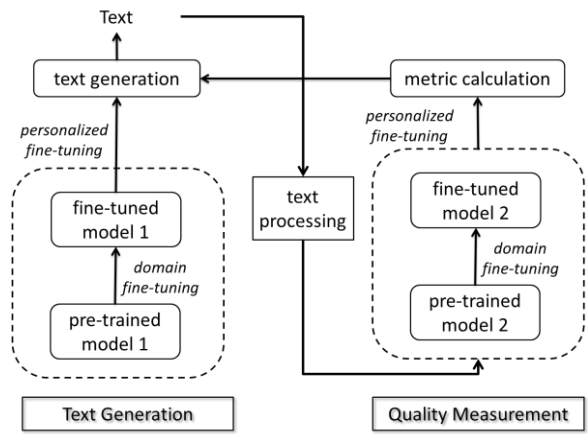

**Fig. 2.** Generic Framework of Text Generation & Measurement

5## 2    Auto-complete function

The auto-complete function is the user interface for inventors to appreciate augmented inventing. This function has four design perspectives: (1) extent of generation, (2) generative direction, (3) proximity of generation, and (4) constraint in generation. Respectively, the extent means how long of the text to generate and measure. It can be a token (sub-word), a word, a phrase, a text span or an entire sentence. The direction means generating text in a forward or backward manner. The proximity means how far the generated text may locate from the current text. The constraint means the requirement to include or exclude a specific text, a type of text or text pattern. The following sections explain these perspectives in details.

### 2.1    Extent of generation

The extent of generation for an inventor depends on the length of text to measure. Different extents of text generation and measurement pose different challenges. In GPT-2, the model generates a token each time by sampling among all tokens in its vocabulary. In fact, the sampling is to measure token probabilities and decide which one to select. Different sampling algorithms mean different ways to select. Most of the language models work at such a token or word level since, by definition, a language model is to predict what the next one token or word is. The reason why expanding the range from "next one" to "next few or many" in this paper is for designing more possible ways of measurement. Technically speaking, the GPT-2 in the framework still generate one token per time, but the BERT model is planned to measure the generated text on a phrase, span or sentence basis. A further assumption is that, in order to measure beyond the token level, a search space of candidate phrases, spans or sentences will be built and explored based on a sequence of tokens generated by GPT-2.

Measuring at the token level is built in GPT-2. A planned experiment is to benchmark performance comparing fine-tuning a pre-trained model by OpenAI and training a model from scratch with patent claims. The token level is the starting point of augmented inventing for inventors, i.e., one word per time. The next level of text generation is to generate phrases. A phrase is a few words with a specific meaning. The phrase could be an inventive element to describe the invention. In order to meet the written requirement mandated by patent law and to avoid ambiguity in patent litigation, it is common to define a key phrase and reuse it elsewhere in patent claims. Such constraint implies that the quality measurement of text generation should be capable of working on a phrase basis. How well the GPT-2 model can meet the written requirement is unknown yet

The next level of text generation is span generation. For human readability, a patent practitioner usually splits a patent claim into several spans. Generally, a patent claim combines several ideas to be innovative. Since the readability in a patent claim aligns with human comprehension of ideas, a claim span becomes a convenient approximation to represent an idea in a patent. My previous experiment in [6] indicates that, at a span level, it is possible to use BERT to measure the relevancy between two spans. Therefore, working on a span basis is an option in terms of measuring text generation. The actual effectiveness of such "a span as an idea" approach remains to be verified. If not compelling enough, a follow-up topic will be how to split a patent claim into more fine-grained or coarse-grained spans. On a span basis, the auto-complete function can provide several generated claim spans for user's selection. Most of patent claims are more prolonged than ordinary sentences. Therefore, text measurement and generation on spans is a middle ground for human comprehension.

The last level of text generation is to generate a complete sentence, i.e., an independent or dependent patent claim. A well-generated independent claim is the combined results of being able to generate the next word, next phrase and next spans well. How to generate a dependent claim is a different challenge because of claim dependency. In patent law, a dependent claim is to describe its corresponding independent claim with specific details. How the framework could learn such claim dependency is to be explored. For example, is it possible to formulate the problem as a Q&A problem by creating an independent claim as a question and its dependent claim as an answer? In this way, the artificial neural network is trained to identify a specific element in the independent claim and generate text to describe the element in details. Section 3.1 will address the issue in training data concerning claim dependency.

## 2.2 Generative direction

After experiments, it was found possible to generate text in a backward manner. By reversing the order of words in input data, the GPT-2 model can be fine-tuned to generate patent claim backwardly. Such backward capability implies that an inventor does not have to draft a patent claim from the beginning. Writing a few words in the middle can develop into both directions. It remains to be seen whether a backward generation can be used to calibrate a forward generation. Another purpose of the backward direction is to explore the possibility of mutual searching between two ideas. For example, the text generation of one claim span can go forward, and the other can go backwards. Is it possible to generate a patent claim incrementally by generating a new claim span to connect two existing claim spans?

## 2.3 Proximity of generation

The proximity means the distance between the current text and the generated text. The reason why this perspective is needed is because of the compositionality in patent claim language. Such compositionality comes from two observations: (i) Technical inventions are generally compositional by combining steps in method/process or different arrangement of components or matters, and (ii) Compositionality is literally codified in patent laws. When drafting patent claims, one can describe and arrange the inventive elements of patent claims in different orders. This indicates that the distance of two inventive elements in prior arts may be different from the order in an inventor's mind. Therefore, it is crucial for the framework to look ahead and provide multiple claim spans at a user's discretion.

It can also be noted that the Transformer models referred to so far work in a Euclidean domain. A Euclidean domain is a domain which has a fixed distance from a given center. For example, the training data for GPT-2 is sequential. The distance between two words in the sequential data is fixed. The distance in attentions to calculate is also fixed. By being not limited to Euclidean, the "proximity" perspective may open a new window to a non-Euclidean domain such as a graph, in which, the distance between words, phrases, spans can be dynamic. In this paper, a hypothesis is that a non-Euclidean approach may fit better the nature of the compositionality in patent claims. Graph-based Transformer model is an emerging field of its own, for example, Koncel-Kedziorski et al. [10] proposed text generation from knowledge graphs with graph Transformers. Whether a graph Transformer can handle the compositionality better is the core issue in my new project granted by the Ministry of Science and Technology (MOST) in Taiwan recently. Therefore, if an issue of a Euclidean domain in this paper could not be solved, there is a chance to solve it in a non-Euclidean domain.

### 2.4 Constraint in generation

Setting a constraint to include or exclude a specific text or text pattern is conceptually rule-based. How to train a neural network to learn a rule-based constraint is still an open challenge. Recently, Keskar et al. [11] propose a conditional Transformer language model for controllable generation. The model is trained to condition on control codes that govern style, content, and task-specific behaviour in text generation. Shen et al. [12] point out that current neural encoder-decoder models conflate both "content selection" and "surface realization" into a black-box architecture. As a result, the content to be described in the text cannot be explicitly controlled. The authors propose a general framework based on variational inference and decoupling content selection from the decoder. Papers like these are pointers for learning and exploring how to control patent claim generation.

In Section 2.1, the necessity of working on a phrase basis was explained regarding the written requirement. Such a requirement is a constraint on patent claim generation. Another example is the antecedent basis in definiteness requirement in patent law. The antecedent basis is a judicially created requirement that stems from written requirement. A claim is indefinite when it contains words or phrases whose meaning is unclear. In practice, if an indefinite article indicates an element, the element as a phrase to be referred by a definite article later should be the same so that definiteness will not be an issue. How well the GPT-2 model can meet the antecedent basis requirement is a research topic in this paper.

## 3 Implementation & Challenge

### 3.1 Data

The first challenge encountered is how to combine independent claims and dependent claims as training data. For example, in **Fig. 1**, the claim 2 depends on the claim 1.

The dependency between claims is a unique issue in patent claim processing, compared with most domains in natural language processing. In other domains, when pre-training or fine-tuning a model, it is common to treat text as a stream of tokens after tokenization. One record in training data may mean one sentence or a paragraph. No dependency exists between two records. The issue with the claim dependency in training data is that, if a record means a dependent claim without including its corresponding independent claim, the context of the dependent claim will be missing. Preferably, the independent claim should be placed in front of its dependent claim so that the model can learn both of them together. However, an independent claim has multiple dependent claims most of the time. Therefore, the model is forced to learn an independent claim repetitively in order to learn its respective multiple dependent claims. Also, an independent claim is longer than its dependent claims usually. If it is longer than the whole length of a training record, the model will not be able to learn the actual part of the dependent claim.

When an independent claim has multiple dependent claims, its dependent claim is likely to be relevant to only a subset of elements in the independent claim. A pragmatic approach is to extract the subset and rewrite the dependent claim to combine the subset as a new independent claim. Such a rewrite algorithm is a challenge of its own. Furthermore, if it is to rewrite all dependent claims as new independent claims, all generated patent claims will be independent claims. A tougher challenge will be how to split generated claims into independent and dependent claims.

At the moment of this writing, two versions of baseline training data are under preparation without any rewrite algorithm. One version is treating a dependent claim as one single training record. The other is prepending the independent claim before a dependent claim to concatenate them as one single record. The purpose is to benchmark and observe the outcome of text generation. In parallel, it is useful for benchmark and test different ways of quality measurement.

### 3.2 Pre-trained Model

How to design and benchmark different models and hyperparameters is the most challenging task in this paper. In my previous work [5] [6], it is convenient to leverage the pre-trained GPT-2 model released by OpenAI. A follow-up research topic is to explore how the fine-tuning result will differ if the GPT-2 model is trained from scratch with patent claims and patent claims only. OpenAI has not released their code for training yet. Therefore, my recent experiment is still a work in progress while trying to learn from different GitHub repositories, hyperparameter settings, tokenization mechanisms, and data structures. The size of the model matter a lot too. The current progress is being able to train the smallest GPT-2 model (117M) from scratch on Google Colab for free. It is planned to test different tokenization mechanisms, such as BPE (Byte Pair Encoding) [13] and SentencePiece [14]. After figuring out suitable experimental settings, it is planned to try bigger GPT-2 models, such as medium (355M), large (774M) and largest (1.5B). The large model is already beyond what Google Colab can run. Luckily I was granted with more than 100 TPUs by the TFRC (TensorFlow Research Cloud) program, which should make it possible to build all sizes of GPT-2 models with patent claims.

### 3.3 Personalization

Section 1.2 explains the inventor-centric data approach in this paper. It is also a hypothesis that fine-tuning a model with inventor-centric data can generate more relevant patent claims to the inventor. Such a hypothesis remains to be validated. A critical issue on validation is how to measure the degree of personalization between the generated patent claims and the original inventor-centric data. One idea is to leverage patent classification as an approximation to measure the relevancy in personalization. For example, the more the classification labels overlap with each other, the higher the personalization relevancy should be. This is also a hypothesis to be validated. In order to calculate the overlap, it is required to predict the classification labels of the generated patent claims. For such a task, it is planned to leverage my previous codebase for patent classification in [4]. Two follow-up topics to explore are: (1) Is classification label a valid approximation to measure personalization? If not, what else? (2) Does the distribution of measurement in a BERT model cover the whole generated text well? Or, does the BERT model measure only a few segments of text instead of the whole?

### 3.4 Fine-tuning a fine-tuned model

A few days after my ai.patent.bot in [5] was online, I observed an interesting text generation which looks partly like a patent and partly like a letter. The suspected root cause is that the email fed into the patent bot contains a more extended email signature with an address. The original GPT-2 model was trained with lots of web data. The web data likely contains emails and addresses. Therefore, a mixed input may make GPT-2 generate a mixed result. This scenario is intriguing and made me recalled that, in the image domain, a weird image could emerge in the transition of transfer learning when learning from lion images to dog images, for example. Can a new patent claim be generated during a transition of transfer learning? Can more of a specific kind of patent claims be generated?

In order to figure it out, an experiment was conducted recently by fine-tuning a fine-tuned model in a "slow-motion" manner. Three stages in the experiment: First, the original GPT-2 model (355M) released by OpenAI is fine-tuned for 6,000 steps with patent claims. Such steps are sufficient to generate text that looks like a patent claim. Second, the original BERT-Base model is fine-tuned with three epochs for patent classification based on CPC section labels (A~H & Y). Third, the fine-tuned GPT-2 model is further fine-tuned with a patent set S1 (belongs to CPC section A but not section G) and fine-tuned slowly to the other patent set S2 (belongs to CPC section G and some belong to section A too). The slow transition from the set S1 to the set S2 is for observing closely by minimal steps in order to catch the fine details in fine-tuning. For example, 512 patent claims are generated after every ten steps of fine-tuning. In terms of observation, the metric is the number of CPC labels measured by BERT. The experiment turned out that, in the transition from set S1 to S2, the total number of label G increases while the total number of label A decreases. Unfortunately, the number of patents with both label A and G does not increase as anticipated. How to generate more patent claims of a specific category by fine-tuning remains a difficult problem.

## 4      Related Work

In the patent field, Aristodemou et al. [15] reviewed 57 recent articles on the use of artificial intelligence methods, machine learning and deep learning approaches for analyzing intellectual property data. Lupu et al. [16] reviewed the state-of-the-art progress on Intellectual Property analytics and pointed out that, among patent-related applications, modern neural networks are applied for machine translation primarily. The authors further anticipated that the remarkable success of deep learning would certainly be tested on patent data someday. To my knowledge, my previous work in [5] is the first to propose patent claim generation and this paper is to push the idea further to personalization.

In the computer science field, the two-stage framework (pre-training & fine-tuning) of Transformer models is so effective that it is declared the arrival of the "ImageNet moment for NLP" [17]. Right after BERT and GPT-2, a variety of Transformer-based models emerged in a relatively short period of time, notably Grover by the University of Washington [18], Transformer-XL [19] and XLNet [20] by CMU and Google, ERNIE 2.0 by Baidu [21], MASS by Microsoft, [22], Evolved Transformer by Google [23], SciBERT by the Allen Institute for Artificial Intelligence [24], VideoBERT by Google [25], DocBERT by the University of Waterloo [26], etc. It is foreseeable that, by the time the prototype of the framework is built in the following months, more advancement in the NLP field and better Transformer models will be available. It should be fruitful to keep an iterative approach and apply state-of-the-art techniques to PatentTransformer.

## 5      Summary

The framework in this paper leverages both of the GPT-2 model and the BERT model. Although these models are state-of-the-art Transformer models, constructing a framework on these building blocks is a new challenge. Applying the framework to the patent corpus is another challenge, due to unique properties in patent claim language and legal requirements in patent law. A further challenge is how to measure and generate personalized patent claims by inventor-centric data and fine-tuning. In the era of artificial intelligence, it is generally accepted that human creativity is what sets humans apart from machines. This paper proposes that "augmented inventing" is a tool to help inventors be more creative in technical inventions.